\documentclass[conference]{IEEEtran}
\IEEEoverridecommandlockouts
\usepackage{cite}
\usepackage{amsmath,amssymb,amsfonts}
\usepackage{algorithm}
\usepackage{algorithmic}
\usepackage{dblfloatfix}
\newcommand{\enrique}[1]{}
\newcommand{\shamay}[1]{}
\DeclareMathOperator{\Exp}{\mathbb{E}}
\usepackage{multirow}
\usepackage{balance}
\usepackage{hyperref}
\usepackage{graphicx}
\usepackage{subcaption}
\usepackage{mwe}
\usepackage{textcomp}
\usepackage{xcolor}
\usepackage{array}
\newcolumntype{y}[1]{>{\let\newline\\\arraybackslash\hspace{0pt}}p{#1}}
\newcolumntype{z}[1]{>{\centering\let\newline\\\arraybackslash\hspace{0pt}}p{#1}}

\def\BibTeX{{\rm B\kern-.05em{\sc i\kern-.025em b}\kern-.08em
    T\kern-.1667em\lower.7ex\hbox{E}\kern-.125emX}}
\begin{document}

\title{Integrated Vehicle Routing and Monte Carlo Scheduling Approach for the Home Service Assignment, Routing, and Scheduling Problem
}

\author{\IEEEauthorblockN{Shamay G. Samuel}
\IEEEauthorblockA{\textit{Department of Computer Science} \\
\textit{Brown University}\\
Providence, USA \\
shamay\_samuel@brown.edu}
\and
\IEEEauthorblockN{Enrique Areyan Viqueira}
\IEEEauthorblockA{\textit{Department of Computer Science} \\
\textit{Brown University}\\
Providence, USA  \\
enrique\_areyanviqueira@brown.edu}
\and
\IEEEauthorblockN{Serdar Kad{\i}o\u{g}lu}
\IEEEauthorblockA{\textit{Department of Computer Science} \\
\textit{Brown University}\\
Providence, USA  \\
serdark@cs.brown.edu}
}

\maketitle

\begin{abstract}
We formulate and solve the H-SARA Problem, a Vehicle Routing and Appointment Scheduling Problem motivated by home services management. We assume that travel times, service durations, and customer cancellations are stochastic. We use a two-stage process that first generates teams and routes using a VRP Solver with optional extensions and then uses an MC Scheduler that determines expected arrival times by teams at customers. We further introduce two different models of cancellation and their associated impacts on routing and scheduling. Finally, we introduce the Route Fracture Metaheuristic that iteratively improves an H-SARA solution by replacing the worst-performing teams. We present insights into the problem and a series of numerical experiments that illustrate properties of the optimal routing, scheduling, and the impact of the Route Fracture Metaheuristic for both models of cancellation.
\end{abstract}

\section{Introduction}
The Home Service Industry deals with the provision of services to people at their homes. Examples of such services include home health care, banking services, and appliance repair services. Due to the rapid current and predicted growth of this industry, the development of computationally efficient and implementable tools is essential to support decision-making. Home services require professional service teams to travel for delivering the services to geographically distributed customers. Service providers often quote an appointment time (planned service start time) to each customer in advance to avoid delivery failure. Therefore, when home service providers plan for service, they need to decide the following: 
\begin{itemize}
    \item The number of service teams to hire (i.e., sizing problem)
    \item The assignment of service teams to the customers (i.e., an assignment problem)
    \item The routing of service teams to customers (i.e., a vehicle routing problem)
    \item The assignment of appointment times for the customers (i.e., appointment scheduling problem). 
\end{itemize}

Additionally, stochasticity is an important property of the Home Service Assignment, Routing, and Appointment Scheduling (H-SARA) problem. our focus is solving H-SARA under three key sources of stochasticity: service duration, travel time, and customer cancellation.

\section{Literature Review}
Based on the breakdown of the H-SARA Problem, we first consider the sizing, assignment, and routing problems. The Vehicle Routing Problem (VRP) is the widely studied generalization of the Traveling Salesman Problem. Given a collection of nodes to visit and a fleet of vehicles, the solution to the VRP is to visit all the nodes exactly once and minimize the total distance traveled by all used vehicles. One variant of the VRP includes vehicle assignment cost considerations, where using a vehicle has an assignment cost associated. This variant combines the sizing, assignment, and routing problems together. 

Another variant of the VRP is the Capacitated VRP (CVRP), where additionally, all nodes have a demand and each vehicle has a capacity. The routes should be generated such that all node demand is met by the vehicles, but the capacities of each vehicle is not exceeded. This variant more accurately models shipping items with larger sizes. In the VRP with Time Windows (VRPTW) variant, each node has to be visited within a certain time frame. This more accurately models nodes available during specific points during the day. Another notable variant is the Stochastic VRP (SVRP) as discussed by Laporte et al., (1992) \cite{svrp} where the travel times and service times are stochastic. Additionally, this variant incorporates an overtime penalty for when the vehicles travel past a certain end time. 

Next we consider the Appointment Scheduling Problem (ASP) literature. The problem of scheduling customers to a single server under stochastic service times is widely studied, and first began with the seminal work of Welch and Bailey (1952) \cite{asp_og}. The goal of the ASP is the creation of a deterministic schedule for customer appointments that minimizes customer waiting time, server idle time and overtime. Other significant early papers include Mercer (1960) \cite{asp_queuing}, who modeled the ASP using tools from Queuing Theory, and Ho and Lau (1992) \cite{asp_simul}, who used numerous simulations.

Many ASP studies are in the environments of outpatient clinics and other appointment-based healthcare systems. The sources of challenge in Healthcare Appointment Scheduling range from the uncertainty in arrival times and service durations, and preferences of the patients and the healthcare providers. Gupta and Denton (2008) \cite{asp_gupta_denton} refer to several complicating factors, including uncertainty in customer arrivals, called ``no-shows", which corresponds to the notion of cancellation in H-SARA.  The overview of Outpatient Appointment literature by Ahmadi-Javid et al., (2017) \cite{outpatient_overview} survey numerous papers with ASP models for various classes of service time distributions and cancellation probabilities. Kong et al., (2013) \cite{asp_cones}, Mak et al., (2015) \cite{asp_lim_dist}, and Kemper et al., (2014) \cite{asp_queue} provide ``distributionally robust" models that depend solely on certain moments of the service time distribution. 

We note that these models often represent the ASP as a $2$-Stage Stochastic Linear Program as first done by Gupta and Denton (2008) \cite{asp_gupta_denton}, with modifications accounting for customer cancellation and preferences. 

Regarding stochasticity in travelling time, we note Kieu et al., (2014) \cite{travel_time} perform a survey of numerous empirical studies of road-based public transport across numerous cities, and classify the travel time distributions based on time of day and spatial breakdowns of bus routes. This breakdown of bus routes is on the section (stop-to-stop), segment (numerous stops), and route (all stops) level. In particular, Cats et al., (2014) \cite{travel_time_stops} determine that section travel time distributions for buses in Stockholm, Sweden take a lognormal distribution.
\section{Problem Definition}

We use bold lowercase symbols $\textbf{x}$ to denote vectors, and calligraphic symbols $\mathcal{X}$ to denote random variables.

The Home Service Assignment, Routing, and Scheduling (H-SARA) Problem takes a set of $N$ nodes representing geographically distributed customers that have to be served within a given day. Associated with H-SARA is an underlying complete graph of customers. We label this graph's nodes using the notation $[N]=\{1, 2, \cdots, N\}$ and additionally consider node $0$ as the origin (i.e., service provider's office). This gives us a distance matrix $d$, where $d(i, j)$ denotes a distance between the pair of nodes $(i, j)$, i.e., the weight of the edge $\{i, j\}$. \enrique{this sounds like all customers need to be served exactly at the start of the day. Is that right? I thought customers were served at different times of the day.} 

\enrique{Let's add more context here, something like: Each customer must be served by a service team... } \shamay{Rephrased scenario better} At the start of each day, each of the $N$ customers requests home service, and the service office must assign teams to provide service to these customers at some point during the day. Each customer has to be served exactly once during the day, and as such is visited by only one service team. We hire $M$ homogeneous service teams, and the value of $M$ is flexible since the service office can hire third-party services for the current day. The cost of hiring any service team $m \in [M]$ is $f_m$. Each service team has standard service hours $[0, L]$. The travel time between any pair of nodes $(i,j)$ is stochastic from a known distribution $\mathcal{D}_{T(i, j)}$ with mean $\mu_{T(i, j)}$ and variance $\sigma_{T(i, j)}^2$. \enrique{The travel cost... } \shamay{Rephrased travel cost} Additionally, this travelling time incurs a unit travel cost of $\lambda_T$ per unit time. For each service team $m$, we must determine their route $\textbf{r}^m$. This is a vector of $n_m$ customer indices denoting the order in which the service team visits them.:
\[\textbf{r}^m=(r^m_0,r^m_1,r^m_2,\cdots, r^m_{n_m},r^m_{n_m+1})\]
\enrique{Notation $r^m_i$ is undefined} \shamay{Introduced customer route indices before}

The route  \enrique{this next sentence feels a bit wordy. Let's just say that routes start and end at the office, i.e., $r^m_0=r^m_{n_m+1}=0$} \shamay{Shortened} The routes $\textbf{r}^m$ start and end at the office, i.e. $r^m_0=r^m_{n_m+1}=0$.

Additionally, we must determine the appointment times for each customer that specifies when their requested service starts. Service only begins when one of the $M$ service teams arrives at the customer's location, and service cannot begin before the customer's specified appointment time. Service times are stochastic from a known distribution $\mathcal{D}_S$ with mean $\mu_S$ and variance $\sigma_S^2$. Finally, each customer has a probability of cancellation of home service $p_C$.

Note that due to stochasticity of service team travel times, service times, and cancellation, there are three possible scenarios:
\begin{itemize}
    
    \item A service team reaches a customer at a time before their appointment time, and has to idle until the scheduled service starts. This incurs an idle cost of $\lambda_I$ per unit time.
    
    \item A service team reaches a customer at a time after their appointment time, and the customer has to wait until the service team arrives and the requested service starts. This incurs a waiting\enrique{waiting?} cost of $\lambda_W$ per unit time.
    
    \item A service team requires additional time past the standard service hours $[0,L]$ to finish servicing all scheduled customers and incurs an overtime cost of $\lambda_O$ per unit time.
    
\end{itemize}

\begin{table*}[t]
\centering
  \begin{tabular}{y{0.18\textwidth}|y{0.18\textwidth}|y{0.56\textwidth}}
    \hline
     \textbf{Parameter} & \textbf{Symbol(s)} & \textbf{Explanation} \\
    \hline
    Number of Customers & $N$ & Number of customers that request home service at the start of the day\\
    End Time & $L$ & End of the standard service hours for all teams $[0,L]$ \\
    Service Time & $\mathcal{D}_S$, $\mu_S$, $\sigma_S$ & Service time distribution with mean $\mu_S$ and variance $\sigma_S^2$  \\
    Mean Travel Speed & $\mu_V$ & Mean travel speed for all teams\\ 
    Travel Time & $\mathcal{D}_{T(i,j)}$, $ \mu_{T(i,j)}$, $ \sigma_{T(i,j)}$ & Distribution of travel time on edge $i\sim j$ with mean $\mu_{T(i,j)}$ and variance $\sigma_{T(i,j)}^2$ \\ 
    Assignment Cost & $f_m$ & Cost of hiring any team $m\in [M]$ \\ 
    Unit Travel Cost & $\lambda_T$ & Cost of service team travel per unit time \\
    Unit Wait Cost & $\lambda_W$ & Cost of customer waiting per unit time \\
    Unit Idle Cost & $\lambda_I$ & Cost of service team idling per unit time \\
    Overtime Factor & $\lambda_O$ & Cost of service team overtime per unit time\\
    Cancel Rate/Probability & $p_C$ & Probability of customer cancellation during the day\\
    \hline
  \end{tabular}
  \caption{H-SARA Parameters}
  \label{tab:hsara_param}
\end{table*}
\begin{table*}[t] 
\centering
  \begin{tabular}{y{0.18\textwidth}|y{0.18\textwidth}|y{0.56\textwidth}}
    \hline
     \textbf{Parameter} & \textbf{Symbol(s)} & \textbf{Explanation} \\
    \hline
    Routing Cost & $\Phi_R$ & Sum of the cost of hiring service teams and the cost of all team travel time\\
    Number of Teams & $M$ & Number of teams to be assigned to serve all customers\\
    Routes & $\{\textbf{r}^m\}_{m=1}^M$ & Set of $M$ vectors of $n_m$ customers representing order service team $m$ visits\\
    \hline
  \end{tabular}
  \caption{Route Assignment Decision Variables}
  \label{tab:hsara_route}
\end{table*}
\begin{table*}[t] 
\centering
  \begin{tabular}{y{0.18\textwidth}|y{0.18\textwidth}|y{0.56\textwidth}}
    \hline
     \textbf{Parameter} & \textbf{Symbol(s)} & \textbf{Explanation} \\
    \hline
    Scheduling Cost & $\Phi_S$ & Sum of customer waiting costs, service team idle costs, and service team overtime costs\\
    Appointment Times & $\{\textbf{a}_m\}_{m=1}^M$ & Set of $M$ vectors of scheduled appointment times for customers on route $\textbf{r}^m$ \\
    Service Team Arrival Time & $\mathcal{T}^m_i$ & Random variable for arrival time of service team $m$ at customer $r^m_i$  \\
    Service Time Variable & $\mathcal{Z}^m_i$ & Random variable for arrival time of service team $m$ at customer $r^m_i$  \\
    Wait Times & $\mathcal{W}^m_i$ & Random variable for waiting time of customer $r^m_i$  \\
    Idle Times & $\mathcal{I}^m_i$ & Random variable for idle time of service team $m$ at customer $r^m_i$  \\
    Overtime & $\mathcal{O}_m$ & Random variable for overtime of service team $m$  \\
    Cancellation Variable & $\mathcal{Y}^m_i$ & Bernoulli$(1-p_C)$ random variable for each customer $r^m_i$. Modifies $\hat{\mathcal{Z}}^m_i=\mathcal{Y}^m_i\mathcal{Z}^m_i$ \\ 
    Cancellation Time & $\mathcal{D}_{C^m_i}$, $\mu_{C^m_i}$, $\sigma_{C^m_i}$ & Cancellation time distribution with support $[0, a^m_i]$, mean $\mu_{C^m_i}$, and variance $\sigma_{C^m_i}^2$\\
    Cancellation Time Variable & $\mathcal{T}_{C^m_i}$ & Cancellation time of customer $r^m_i$\\
    \hline
  \end{tabular}
  \caption{Appointment Scheduling Decision Variables}
  \label{tab:hsara_sched}
  \enrique{this looks amazing! :-)}
\end{table*}

The solution to an H-SARA problem instance consists of routing assignments and customer schedules. The quality of an H-SARA solution is a function of the total service team assignment cost, traveling cost, service team idling cost, customer waiting cost, and overtime cost. To construct the objective function $\Phi$ representing solution quality, we first split the categories of cost into two types:
\begin{enumerate}
    \item \textbf{Routing Cost}: The Routing Cost, $\Phi_R$, \enrique{could we write the domain and range of $\Phi_R$?} \shamay{Added domain and range. Not sure about how I can represent the space of all possible routes as a set :(. Maybe I could use some notation for set partitions of $[N]^M$} \enrique{yeah, now I think it might be better to omit the domain. A proper solution probably involves a more careful definition of the domain, so let's ignore that for this friday's deadline} is the sum of the cost of service team assignment and cost travel time:
    \begin{equation}
        \Phi_R:=Mf_m + \lambda_T\sum_{m=1}^M\sum_{i=0}^{n_m} t(r^m_i,r^m_{i+1})
    \end{equation}

    Where $t(i,j)$ is a non-negative random variable denoting the travel time of the vehicle between customers pair $(i,j)$. Letting $v(i,j)$ be the velocity of the vehicle between customers $(i,j)$, we get:
    \begin{equation}
        \Phi_R:=Mf_m + \lambda_T\sum_{m=1}^M\sum_{i=0}^{n_m} \frac{d(r^m_i,r^m_{i+1})}{v(r^m_i,r^m_{i+1})}
    \end{equation}
    Minimization of Routing Cost is discussed in Section \hyperref[sec:routing]{\color{blue}VI}.
    
    \item \textbf{Scheduling Cost}: The Scheduling Cost, $\Phi_S$, \enrique{could we write the domain and range of $\Phi_S$?} \shamay{Similar issue with inelegant set notation}\enrique{similar comment as above} is the sum of the wait costs, idle costs, and overtime costs. Consider a service team $m$ with route $\textbf{r}^m$. We define the deterministic vector $\textbf{a}^m$ for the appointment times provided by the office to the customers in $\textbf{r}^m$ at the start of the day, and the random vector $\textbf{t}^m$ representing service team arrival times at customers in $\textbf{r}^m$:\enrique{I think $\textbf{t}$ should be lowercase but the coordinates are fine as they are}
    \[\begin{split}
        \textbf{a}^m&=(a^m_0,a^m_1,\cdots, a^m_{n_m}, a^m_{n_m+1})\\
    \textbf{t}^m&=(\mathcal{T}^m_0,\mathcal{T}^m_1,\cdots, \mathcal{T}^m_{n_m}, \mathcal{T}^m_{n_m+1})
    \end{split}\]
    \enrique{is the double sub-index necessary here $ a^m_{n_m}$? is hard to parse...} \shamay{Introduced $n_m$  instead of $n_m$ that specifies the length of $\textbf{r}^m$ instead} \enrique{still hard to parse, but let's move on for friday's submission}
    For each customer $r^m_i\in[N]$ on the route, we have their scheduled appointment time $a^m_i$ and the actual arrival time of the service team $\mathcal{T}^m_i$. As such, we set $a^m_0=\mathcal{T}^m_0=0$ as the time each service team starts at the office.
    
    For the sake of simplicity, we introduce the inter-schedule vector $\textbf{x}^m=(x^m_1,\cdots, x^m_{n_m}, x^m_{n_m+1})$ denoting the sequence of inter-schedule times for service team $m$ on $\textbf{r}^m$, i.e. $x^m_i=a^m_i-a^m_{i-1}$. 
    We then let $\mathcal{Z}^m_i\sim \mathcal{D}_S$ be a random variable denoting a random service time for customer $r^m_i$. 
    \enrique{suggestion: let's decorate random variables, e.g., $\mathcal{Z}$, so that the reader knows what is random and what is deterministic. For example, when I was reading it took me a bit to get that $W$ is actually random, but we can make this notation easier by writing $\mathcal{W}$. Also, if you do follow this suggestion, let the reader know at the beginning of the definition section that you'll use calligraphy for r.v.s, lower case for vectors, etc...} 
    
    We can now construct a recursive form similar to Gupta and Denton (2008) \cite{asp_gupta_denton} for wait times $\mathcal{W}^m_i$ and idle times $\mathcal{I}^m_i$ for the customers in $\textbf{r}^m$, and the overtime $\mathcal{O}_m$ for the service team $m$:
    \begin{equation} \label{eqn:sched}
        \begin{split}
    \mathcal{W}^m_{i+1}&=(\mathcal{W}^m_{i}+\mathcal{Z}^m_{i}+t(r^m_i,r^m_{i+1})-x^m_{i+1})^+\\
    \mathcal{I}^m_{i+1}&=(x^m_{i+1}-\mathcal{W}^m_{i}-\mathcal{Z}^m_{i}-t(r^m_i,r^m_{i+1}))^+\\
        \mathcal{O}_m&=(\mathcal{W}^m_{n_m}+\mathcal{Z}^m_{n_m}+a^m_{n_m}+t(r^m_{n_m},r^m_{n_m+1})-L)^+\\
        &=\left(\mathcal{W}^m_{n_m}+\mathcal{Z}^m_{n_m}+\sum_{i=1}^{n_m}x^m_i+t(r^m_{n_m},0)-L\right)^+
    \end{split}
    \end{equation}
    Where $(\cdot)^+$ is the positive part function, i.e. $\max(\cdot, 0)$. This gives us the Scheduling Cost for each service team $m$ as follows:
    \begin{equation}
        \Phi^m_S:=\lambda_W\sum_{i=1}^{n_m} \mathcal{W}^m_i +\lambda_I\sum_{i=1}^{n_m} \mathcal{I}^m_i +\lambda_O \mathcal{O}_m 
    \end{equation}
    Summing over all service teams $m\in[M]$, we get the Scheduling Cost:
    \begin{equation}
        \Phi_S:=\sum_{m=1}^M \Phi^m_S
    \end{equation}
    Minimization of Scheduling Cost is discussed in Section \hyperref[sec:scheduling]{\color{blue}VII}.
\end{enumerate}
Combining both of the above costs, we obtain the following objective function representing Total Cost:
\begin{equation}
    \Phi:=\Phi_R + \Phi_S
\end{equation}
\enrique{what are these weights? are these parameters? do we chose them?} \shamay{specified the role of these weights} \enrique{If we never try different values for these weights, I suggest get rid of them and save the reader cognitive power for the rest of the model. No point in learning about weights that are never used.} \shamay{removed weights} Our goal is to minimize $\Phi$ in expectation.
\ref{tab:hsara_sched}. 
\begin{figure*}[t]
    \centering
    \begin{subfigure}[b]{0.3\linewidth}
        \centering
        \includegraphics[width=\linewidth]{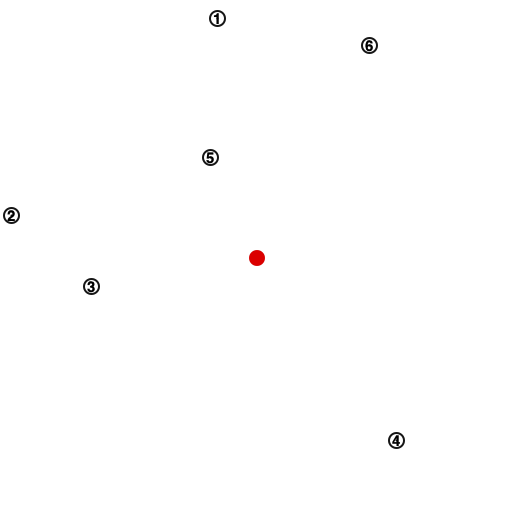}
        \caption[Routes]{{\small $N=6$ Customers}}    
        \label{fig:example_custs}
    \end{subfigure}
    \begin{subfigure}[b]{0.3\linewidth}  
        \centering 
        \includegraphics[width=\linewidth]{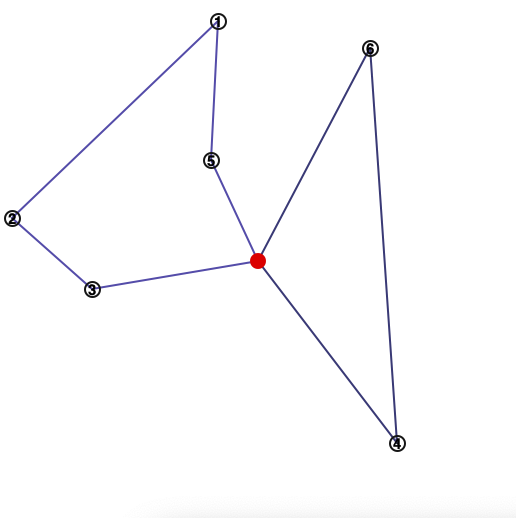}
        \caption[Routes]{{\small Routes for $M=2$ Service Teams}}    
        \label{fig:example_routes}
    \end{subfigure}
    \begin{subfigure}[b]{0.3\linewidth}   
        \centering 
        \includegraphics[width=\linewidth]{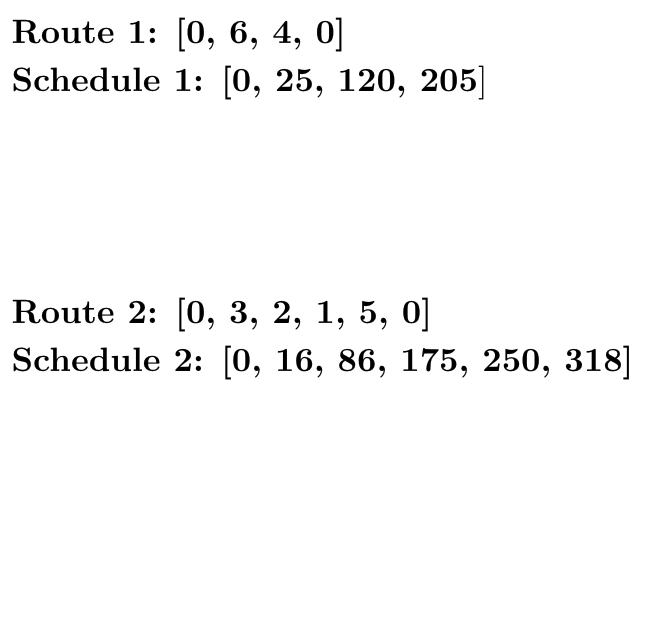}
        \caption[Routes]{{\small Schedules for $M=2$ Service Teams}}    
        \label{fig:example_sched}
    \end{subfigure}
    \caption{Example H-SARA Problem Customer Locations, Routes, and Schedules}
    \label{fig:examples}
\end{figure*}
We now address customer cancellation via two different models of cancellation:
\begin{enumerate}
    \item \textbf{Last-Minute Cancellation:} This model makes the assumption that the service teams \enrique{is it important that both service teams AND office does not know about cancellations? Are there separate roles for the office and service teams in the model?} do not know about customer cancellation until they visit the customer. This model represents last-minute cancellations by customers or ``forgetful" customers. Following this model, we can modify the random service time variable $\mathcal{Z}^m_i$. Let $\mathcal{Y}^m_i$ be a Bernoulli$(1-p_C)$ \enrique{I realize now there are a lot of letters in this model. Would it be possible to make a two-column table, left column the name of the parameter (e.g., $p_C$, and right column a brief explanation (e.g., cancellation rate)?} random variable independent of $\mathcal{Z}^m_i$, and let $\hat{\mathcal{Z}}^m_i=\mathcal{Y}^m_i\mathcal{Z}^m_i$. $\hat{\mathcal{Z}}^m_i$ denotes the service time at customer $r^m_i$ with cancellation probability $p_C$, that takes the value of $\mathcal{Z}^m_i$ with probability $1-p_C$ and zero otherwise. 
    
    We now only need to modify $\Phi_S$, since the service teams will have to travel to the customers to learn about cancellation. In this case, customer wait times, service team idle times, and service team overtime now depend on $\hat{\mathcal{Z}}^m_i$ instead of $\mathcal{Z}^m_i$. We denote the H-SARA Problem that uses this model of cancellation as H-SARA-0.
    
    \item \textbf{Notified Cancellation:} This model makes the assumption that customers notify the office of cancellation at some point before their appointment time. This model represents more ``responsible" customers who notify the office of their service cancellation in advance. A customer $r^m_i$'s cancellation time $\mathcal{T}_{C^m_i}$ follows a known distribution $\mathcal{D}_{C^m_i}$ with support $[0, a^m_i]$, mean $\mu_{C^m_i}$, and variance $\sigma_{C^m_i}^2$. Based on the notification time of cancellation \enrique{notification time of cancellation sounds funny but I couldn't think of something better...}, we allow service teams to reroute to potentially minimize travel times and their Scheduling Cost. We denote the H-SARA Problem that uses this model of cancellation as H-SARA-1.
\end{enumerate}
The naming of these problem variants is chosen such that we treat the models as two extremes and can generalize them \enrique{Might save space to just explain H-SARA-$\lambda$, and then instantiate the two models $\lambda=0$ and $\lambda=1$, along with a brief description of them} into H-SARA-$\lambda$. Here $\lambda\in[0,1]$ and denotes the probability that a customer who cancels notifies the office in advance. When $\lambda=0$, we have that no customer notifies the office, giving us the first model of cancellation. When $\lambda=1$, we have that every customer notifies the office prior to their appointment time, giving us the second model. These models of cancellation are further discussed in Section \hyperref[sec:cancellation]{\color{blue}VIII}.

A summary of all the parameters in this model is provided in Tables \ref{tab:hsara_param}. The decision variables, associated variables, and distributions are provided in  \ref{tab:hsara_route} for the Routing Assignment, and \ref{tab:hsara_sched} for the Appointment Scheduling.


\subsection{Illustrative Example} 
Let us make the abstract problem definition more concrete with an illustrative example. 

As shown in Figure \ref{fig:example_custs}, we have 6 customers geographically distributed around a depot. \enrique{looks like we have more customers! also, Fig 1 is so far down, maybe say... As shown in Figure 1 (page x)...} Further suppose that the H-SARA Parameters are as follows:

\noindent\makebox[\linewidth]{\rule{\linewidth}{1pt}}
\small $L = 480$ (Standard Service Hours)\\
$f_m=100$ (Service Team Assignment Cost)\\
$\lambda_T=1.5$ (Unit Travel Cost)\\
$\lambda_W=10$ (Unit Customer Waiting Cost)\\
$\lambda_I = 5$ (Unit Service Team Idle Cost)\\
$\lambda_O = 15$ (Unit Service Team Overtime Cost)\\
$p_C= 0.1$ (Customer Service Cancellation Rate)\\
\noindent\makebox[\linewidth]{\rule{\linewidth}{1pt}}
\normalsize
Now suppose we determine that $M=2$ is the required number of service teams, and the required routes and schedules are as in Figure \ref{fig:example_sched}. This gives us Service Assignment Team Cost $\Phi_A$:
\[\Phi_A = Mf_m = 2\cdot 100 \]
Now the Travel Cost $\Phi_{T}^m$ for each route $\textbf{r}^m$:
\[\begin{split}
    \Phi_{T}^1 &= \lambda_T\sum_{i=0}^{n_1} t(r^1_i,r^1_{i+1}) \\ &= 1.5\left(t(\textbf{0},\textbf{6})+t(\textbf{6},\textbf{4})+t(\textbf{4},\textbf{0})\right)\\
    \Phi_{T}^2 &= \lambda_T\sum_{i=0}^{n_2} t(r^2_i,r^2_{i+1})\\  &= 1.5\left(t(\textbf{0},\textbf{3})+t(\textbf{3},\textbf{2})+t(\textbf{2},\textbf{1})+t(\textbf{1},\textbf{5})+t(\textbf{5},\textbf{0})\right)
\end{split}\]
This gives us the Routing Cost as:
\[\Phi_R=\Phi_A + \Phi_{T}^1 + \Phi_{T}^2\]
Now expressing the Scheduling Cost $\Phi_S$ first requires the inter-schedule times for each route:
\[\begin{split}
\textbf{x}^1&=(25, 95, 85)\\
\textbf{x}^2&=(16, 70, 89, 75, 68)\end{split}\]
The Customer Waiting times, Service Team Idle times and Service Team Overtime can be constructed using the recurrence in Equation \ref{eqn:sched}. This gives us the Customer Waiting Cost, Service Team Idle Cost, and Service Team Overtime cost for the first team:
\[\begin{split}
    \Phi^1_W&=\lambda_W\sum_{i=1}^{n_1} \mathcal{W}^1_i = 10\sum_{i=1}^{n_1} \mathcal{W}^1_i\\
    \Phi^1_S&=\lambda_I\sum_{i=1}^{n_1} \mathcal{I}^1_i =5\sum_{i=1}^{n_1} \mathcal{I}^1_i\\
    \Phi^1_O &= \lambda_O \mathcal{O}_1 = 15\mathcal{O}_1
\end{split}
\rightarrow
\begin{split}
    \Phi^1_S = \Phi^1_W + \Phi^1_I + \Phi^1_O
\end{split}
\]
Similarly, for the second team:
\[
\begin{split}
    \Phi^2_W&=\lambda_W\sum_{i=1}^{n_2} \mathcal{W}^2_i= 10\sum_{i=1}^{n_2} \mathcal{W}^2_i\\
    \Phi^2_S&=\lambda_I\sum_{i=1}^{n_2} \mathcal{I}^2_i =5\sum_{i=1}^{n_2} \mathcal{I}^2_i\\
    \Phi^2_O &= \lambda_O \mathcal{O}_2 = 15\mathcal{O}_2
\end{split}
\rightarrow
\begin{split}
    \Phi^2_S = \Phi^2_W + \Phi^2_I + \Phi^2_O
\end{split}\]
Combining the Scheduling Costs for both teams gives us the Scheduling Cost:
\[\Phi_S= \Phi^1_S+\Phi^2_S\]
Therefore, we can now express the Total Cost for this H-SARA instance as:
\[\Phi = \Phi_R + \Phi_S\]
The value we seek to minimize is the expected Total Cost:
\[\begin{split}
    \Exp[\Phi]&=\Exp[\Phi_R] + \Exp[\Phi_S]\\
    &=\Phi_A + \sum_{m=1}^2\Exp_{t\sim\mathcal{D}_{T}}[\Phi_{\textbf{r}^m}] + \Exp_{t\sim\mathcal{D}_{T}, \mathcal{Z}\sim \mathcal{D}_S}[\Phi^m_{T}]
\end{split}\]
The Customer Cancellation Rate $p_C=0.1$ modifies the Total Cost based on the model of cancellation:
\begin{enumerate}
    \item In H-SARA-$0$, for each route $m$, we create random variables $\mathcal{Y}^m_i\sim$Bernoulli$(0.9)$ that modify the service times $\mathcal{Z}^m_i$ in Equation \ref{eqn:sched} to $\hat{\mathcal{Z}^m_i}=\mathcal{Y}^m_i\mathcal{Z}^m_i$. This changes the Scheduling Cost, but not the Routing Cost.
    \item In H-SARA-$1$ , for each route $m$, customer $r^m_i$ cancels at time $\mathcal{T}_{C^m_i}$, and we allow the teams to reroute and skip a canceled customer if we predict rerouting would reduce the team's Routing and Scheduling Costs. 
\end{enumerate}
\section{Case Study}
In this section, we introduce the Case Study for H-SARA and our overall design for producing a near-optimal solution to it. We solve H-SARA problem instances with $N = 20$, $30$, $40$ and $50$ customers. We have the origin node representing the office at $(0, 0)$ and the customers are distributed uniformly in a square of edge $50$ kilometers with the origin at the center. For each instance, we first generate N nodes $(a_i ,b_i)$ in this region then compute the distance between each pair of customer nodes $(i,j)$ using the Euclidean metric on $\mathbb{R}^2$:
\[d(i,j)=\sqrt{(a_i-a_j)^2+(b_i-b_j)^2}\]
The mean value travel time $t(i,j)=d(i,j)/v(i,j)$, where $v(i,j)$ represents the mean travel velocity between the pair of nodes $(i,j)$, and its value is equal to $1$km/minute in all instances. 

The cost of assigning one service team is generated from $U[100, 250]$, where $U[a,b]$ denotes the
uniform distribution on $[a,b]$.\enrique{is $U[a,b]$ defined somewhere?} \shamay{Added $U[a,b]$ definition} We set the unit travel time cost $\lambda_T\in \{0.5,1,2\}$. The mean customer’s service time $\mu_S$ is generated from $U[30,60]$ minutes, and the standard deviation of service time is set to $0.5\mu_S$. We also generate the customer's Unit Wait Cost $\lambda_W$ from $U[0,10]$, the service team's Unit Idle Cost $\lambda_I$ from $U[0,5]$, and the service team overtime unit cost $\lambda_O$ from $U[1.5, 10]$. Finally, \enrique{too many use of 'We'...} \shamay{Attempted a fix} problem instances are generated with cancellation probability $p_C\in\{0.01, 0.05, 0.1\}$.

\section{Solving H-SARA}
\enrique{unclear what 'this' is referring to. I would get rid of the previous sentences. More broadly, try to avoid starting a sentence with 'this' because it often confuses what the object of the sentences if referring to} \shamay{Deleted sentence} In order to solve the Case Study for the H-SARA Problem, we model it as a two-stage process. The first stage is formulating the service team number and route assignment problems as a Vehicle Routing Problem with Fixed Vehicle Costs. Upon generating routes, we then determine customer appointment times using a series of Monte Carlo simulations that updates a base schedule with the appropriate offsets for idling times. \enrique{this paragraph is very important, I feel it should be either highlighted or at the beginning of this section}. \shamay{I set it as a new section}

Finally, after generating a list of the best potential solutions we incorporate a metaheuristic that, for each of the solutions, reroutes and reschedules the service teams with the worst simulated total costs.
\section{Route Assignment}
\label{sec:routing}
We now develop the techniques used to solve H-SARA's Route Assignment. In this stage, we must determine the number of service teams $M$ and generate service team routes. We model this problem as a Vehicle Routing Problem (VRP). 

Let $G = (V,E)$ be the complete graph on $|V|=N+1$ vertices indexed as $\{0,1,\cdots, N\}$, and $E=\{d(i,j)\;|\;i,j\in V\}$ \enrique{Sometimes you write $d_{i,j}$, sometimes $d(i,j)$. Is there a reason for these two different notations?} \shamay{fixed} is the set of weighted edges representing distances between the vertices (i.e. customers). We let the vertex with index $0$ denote the depot (i.e. service provider's office) located at the origin. We further have $M$ vehicles (i.e. service teams) and the cost of assignment for each vehicle is $f_m$. The goal of the VRP is to generate routes $\textbf{r}^m$, for each vehicle $m$, such that the total travel distance for all vehicles is minimized:\enrique{I think you want to write the $\min$ operator here, right? as in, the optimization problem.}
\[\Phi_D = \min\left\{\sum_{m=1}^M\sum_{i=0}^{n_m-1} d(r^m_i,r^m_{i+1})\right\}\]
For the case study, since for any pair of customers $(i,j)$ we have $t(i,j) \sim \mathcal{D}_{T(i,j)}$ which has mean $d(i,j)$, minimizing the total travel distance for all vehicles minimizes the total travel time for all vehicles in expectation.\enrique{not sure about this last sentence. I think you want to write the optimization problem as minimizing the travel distance in expectation where travel times are drawn from the distribution you mentioned. Otherwise, the optimization problem is not well-defined and this last sentences is confusing. } \shamay{fixin this}

Our task is to consider different numbers of vehicles $M$ and generate routes for the vehicles such that each customer is visited exactly once and the total travel distance $\Phi_D$ is minimized. We then choose the value of $M$\enrique{best/worst values of $M$ is undefined. we don't have a measure of the cost of number of vehicles, do we? this number is dependent on the routes themselves, but you make it sounds as is the problem decomposes in two stages (it does not, we are decomposing it that way heuristically)} such that the Routing Cost $\Phi_R$ is minimized. We further consider the following extensions to generate additional solutions.

\subsection{Vehicle Capacities} 
The Capacitated Vehicle Routing Problem (CVRP) incorporates vehicle capacities and customer demands as additional constraints. Each vehicle $m\in M$ has limited capacity $L_m$ and each customer $r^m_i$ has a demand $w^m_i$. Solving the CVRP generates routes for $M$ vehicles such that for each route $\textbf{r}^m$, the route satisfies:
\[\sum_{i=1}^{n_m}w^m_i\leq L_m\]
To model the H-SARA Route Generation problem as a CVRP, we treat the vehicle capacity as the end of the standard service hours $L$, and the demands of each customer as the expected service time with cancellation $(1-p_C)\mu_S$. Note that the total demand of all customers is $N(1-p_C)\mu_S$, so we have:
\[N(1-p_C)\mu_S\leq ML\implies M \geq \left\lceil\frac{N(1-p_C)\mu_S}{L}\right\rceil\]
\enrique{time permitting, I suggest you add an example here}
\subsection{Customer Time Windows} 
The Vehicle Routing Problem with Time Windows (VRPTW) incorporates a time window for each customer indicating when the visit should be made. Each customer $i$ has a time window $(t_i, t_i')$ when a vehicle should visit. Further, the depot has operational hours $(0, t_D)$ representing when each vehicle starts and when each vehicle should return by. Another assumption is that the service time at each customer is $0$.

To model the H-SARA Route Generation problem as a VRPTW, we specify a time window of $(0,L)$ for each customer and the office.\enrique{i don't really understand this last sentence... }

\section{Appointment Scheduling}
\label{sec:scheduling}
After generating the routes for $M$ service teams, we then move to the second stage of Appointment Scheduling for the H-SARA problem. In this stage, we create a baseline appointment schedule and improve it using a Monte Carlo Simulator to more accurately predict the arrival times for a service team $m$ at the customers $r^m_i$ on its route $\textbf{r}^m$.
\subsection{Fixed Schedule}
We begin by defining the Baseline Appointment Schedule $\textbf{B}^m$ for the customers in $\textbf{r}^m$ as:
\[B^m_0= 0,\qquad B^m_1 = d(0, r^m_1)\]
\[B^m_i=B^m_{i-1}+(1-p_C)\mu_s + d(r^m_{i-1}, r^m_i)\]
In other words, we treat the service times as deterministic with value $(1-p_C)\mu_S$, i.e. the expected service time at each customer with cancellation, and add the expected travel time $\Exp[t(r^m_{i-1},r^m_i)]=d(r^m_{i-1},r^m_i)$ between each customer on the route. 
\subsection{Simulated Schedule}
We now describe a Monte Carlo Scheduler. The goal is to update a given schedule to more accurately reflect a service team's actual arrival times to their assigned $n_m$ customers. Consider the case of H-SARA-$0$, or for H-SARA-$1$ with $p_C=0$. For a service team $m$ with route $\textbf{r}^m$ and schedule $\textbf{a}^m$, the arrival time $\mathcal{T}^m_i$ at customer $r^m_i$ takes the following recursive form\enrique{let's try to break this last sentence into 2 or more sentences}:
\[\mathcal{T}^m_i = \max(\mathcal{T}^m_{i-1}, a^m_{i-1}) + \hat{\mathcal{Z}}^m_{i-1}+t(r^m_{i-1},r^m_i)\]
This gives us:
\[\Exp[\mathcal{T}^m_i] = \Exp[\max(\mathcal{T}^m_{i-1}, a^m_{i-1})] + (1-p_C)\mu_S+d(r^m_{i-1},r^m_i)\]
So we set the arrival time at the office as $\mathcal{T}^m_{0}=0$, and can use Monte Carlo simulation to estimate the expected arrival times at all customers on $\textbf{r}^m$:
\begin{algorithm}[H]
\caption{Monte Carlo Scheduler}
\begin{algorithmic}[1]
\renewcommand{\algorithmicrequire}{\textbf{Input:}}
\REQUIRE $\{\textbf{r}^m\}_{m=1}^M$, $\{\textbf{a}^m\}_{m=1}^M$, $N_{\text{iter}}$
\FOR {$n=1$ to $N_{\text{iter}}$}
\FOR {$m = 1$ to $M$}
\STATE $a_{\text{new}}^m\leftarrow\Exp$[SimulateArrivalTimes$\left(\textbf{r}^m, \textbf{a}^m\right)$]
\enrique{when writing the expectation operator, please write what is the expectation being taken over as a subscript}
\ENDFOR
\STATE $\{\textbf{a}^m\}_{m=1}^M\leftarrow\{\textbf{a}_{\text{new}}^m\}_{m=1}^M$
\ENDFOR
\RETURN $\{\textbf{a}^m\}$
\end{algorithmic}
\end{algorithm}
The description of SimulateArrivalTimes is available in Appendix \hyperref[adx:misc]{A}. In the case of H-SARA-$1$ with $p_C>0$, the recursive form for $\mathcal{T}^m_i$ \enrique{label the equations (you can use the equation environment) and reference equations by number. Avoid saying the above, is confusing}  does not necessarily hold, since the service team can choose to dynamically reroute based on the notification of customer cancellation. This results in modification to the route $\textbf{r}^m$ during the simulation and prevents a clean closed form representation of arrival times. However, we can still empirically estimate the arrival times at each customer using the above scheduler.
\section{Cancellation Models}
\label{sec:cancellation}
In this section, we explore the two different models of cancellation based on how service teams are notified, and the impact they have on the route generation and appointment scheduling stages of our solution:

\subsection{Last-Minute Cancellation (H-SARA-0)}
In this model, customer cancellations are only learned when their assigned service team visits them at some point during the day. Note that this model only affects the Scheduling Cost, since service teams have to travel to the customer's location to learn about cancellation. This results in the following considerations during the modelling of the two-stages:
\begin{itemize}
    \item During service team and route assignment, when we consider the CVRP model of route generation, we specifically incorporate the probability of cancellation into the capacities of vehicles. For $M$ vehicles, we obtain routes such that for each route $\textbf{r}^m$, the route satisfies:
    \[\sum_{i=0}^{n_m-1}w^m_i\leq L_m\]
    Where $w^m_i$ denotes the demand of each customer $r^m_i$ on the route, and $L_m$ denotes the capacity of vehicle $m$. We now treat the  service time $\hat{\mathcal{Z}}^m_i$ as the demands of the customer $r^m_i$, and the capacity of each service team $m$ as the end of the standard service hours $L$. Using the fact that $\mathcal{Y}^m_i,\mathcal{Z}^m_i$ are independent, this gives us the deterministic weights as follows:
    \[w^m_i:=\Exp[\hat{\mathcal{Z}}^m_i] = \Exp[\mathcal{Y}^m_i \mathcal{Z}^m_i]=(1-p_C)\mu_S\]
    \item During appointment scheduling for a route $\textbf{r}^m$, we establish a baseline appointment schedule $\textbf{B}^m$, by treating the customer services times as deterministic with value $\Exp[\hat{\mathcal{Z}}^m_i]=(1-p_C)\mu_S$ and adding the expected travel time $\Exp[t(r^m_{i-1},r^m_i)]=d(r^m_{i-1},r^m_i)$ between each customer on the route. 
    
    Further, when we use the Monte Carlo Scheduler to obtain a simulated schedule for H-SARA-$0$, and H-SARA-$1$ with $p_C = 0$, we obtain the following recurrence form for the expected arrival time of service team $m$ at customer $r^m_i$:
    \[\Exp[\mathcal{T}^m_i] = \Exp[\max(\mathcal{T}^m_{i-1}, a^m_{i-1})] + (1-p_C)\mu_S+d(r^m_{i-1},r^m_i)\]
\end{itemize}

\subsection{Notified Cancellation (H-SARA-1)}
In this model, the cancellation by a customer are learned by their assigned service team at a random time before the customer's scheduled appointment time. In particular, a customer $r^m_i$'s cancellation time follows a known distribution $\mathcal{D}_{C^m_i}$ with support $[0, a^m_i]$, mean $\mu_{C^m_i}$, and variance $\sigma_{C^m_i}^2$. These cancellation distributions $\mathcal{D}_{C^m_i}$ allow us to more flexibly reflect different styles of cancellations, ranging from uniform to ``earlier" and ``later" cancellations based on closeness to the appointment time.

We further allow in this model a service team $m$ to reroute upon being notified of a cancellation of a customer $r^m_i$ to potentially reduce the simulated total cost. We use the following simple heuristic during the simulation of the service team $m$'s route:

\begin{algorithm}[H]
\caption{H-SARA-1 Reroute Heuristic}
\begin{algorithmic}[1]
\renewcommand{\algorithmicrequire}{\textbf{Input:}}
\REQUIRE $r^m_i, r^m_{i+1}, r^m_{i+2}$, $a^m_i, a^m_{i+1}, a^m_{i+2}$
\FOR {$i = 0$ to $n_m-1$}
\IF {$r^m_{i+1}$ is Cancelled}
\STATE $T_1\leftarrow \mathcal{T}^m_{i} + d(r^m_{i},r^m_{i+1}) + d(r^m_{i+1},r^m_{i+2})$
\STATE $T_2\leftarrow \mathcal{T}^m_{i} + d(r^m_{i},r^m_{i+2})$
\IF {SimulateCosts$(T_2) <$ SimulateCosts$(T_1)$}
\STATE $r^m_{i+1}\leftarrow r^m_{i+2}$ 
\STATE $a^m_{i+1}\leftarrow a^m_{i+2}$ 
\ENDIF
\ENDIF
\ENDFOR
\end{algorithmic}
\end{algorithm}
The description of SimulateCosts is available in Appendix A. To simplify notation, we introduce the following variables:
\[R_1=\lambda_T(d(r^m_{i},r^m_{i+1})+d(r^m_{i+1},r^m_{i+2}))\]
\[S_1=\lambda_W(T_1-a^m_{i+2})^++\lambda_I(a^m_{i+2}-T_1)^+\]
\[R_2=\lambda_Td(r^m_{i},r^m_{i+2})\]
\[S_2=\lambda_W(T_2-a^m_{i+2})^++\lambda_I(a^m_{i+2}-T_2)^+\]
The above heuristic allows the service team $m$ to skip visiting a canceled customer $r^m_{i+1}$, and reroute to $r^m_{i+2}$ from $r^m_{i}$ if the following inequality holds:
\[R_2+S_2 < R_1+S_1\]
This heuristic allows for rerouting and minor cost optimizations during the simulation and actively uses information about customer cancellation to its advantage.
\section{Route Fracture Metaheuristic}
\label{sec:metaheuristic}
The above two-stage process generates batch of routes for varying values of $M$ and different VRP variants, and then generates appointment times for each route. We then use the following algorithm to improve each solution:
\begin{figure*}[!t]
\centering
    \begin{subfigure}[b]{0.475\linewidth}
        \centering
        \includegraphics[width=\linewidth]{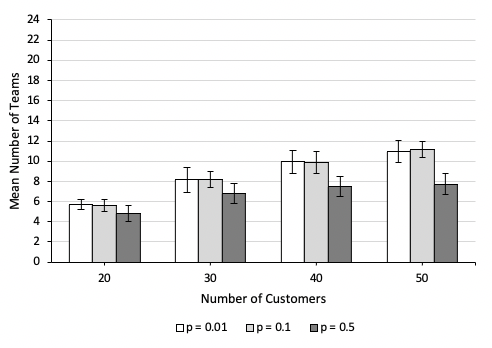}
        \caption[Routes]{{\small No RF Metaheuristic}}    
        \label{fig:M_nometa}
    \end{subfigure}
    \begin{subfigure}[b]{0.475\linewidth}  
        \centering 
        \includegraphics[width=\linewidth]{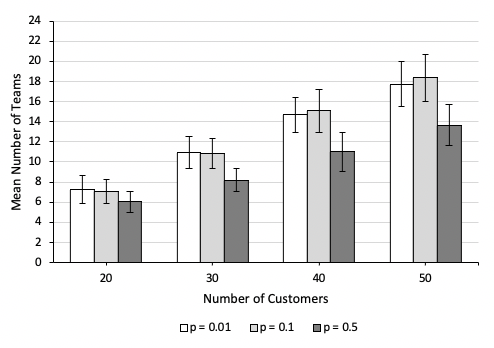}
        \caption[Routes]{{\small RF Metaheuristic}}    
        \label{fig:M_meta}
    \end{subfigure}
    \caption{\small Mean Value for Number of Service Teams $M$ assigned in generated H-SARA solutions}
    \label{fig:M_vals}
\end{figure*}
 \begin{algorithm}[H]
 \caption{Route Fracture Algorithm}
 \begin{algorithmic}[1]
 \renewcommand{\algorithmicrequire}{\textbf{Input:}}
 \REQUIRE $\{\textbf{r}^m\}_{m=1}^M$, $\{\textbf{a}^m\}_{m=1}^M$
  \STATE SimulateCosts$\left(\{\textbf{r}^m\}_{m=1}^M, \{\textbf{a}^m\}_{m=1}^M\right)$
  \FOR {$i = 1$ to $M$}
  \STATE $\text{WT}_i\leftarrow$ WorstSimulatedTeams($i$)
  \STATE $\text{Loc}_i\leftarrow \bigcup_{\textbf{r}^m\in \text{WT}_i}\bigcup_{r^m_j\in \textbf{r}^m} r^m_j$
  \STATE $\{\textbf{r}^{m'}\}_{m'=1}^{M'}, \{\textbf{a}^{m'}\}_{m'=1}^{M'}\leftarrow$ H-SARA$(\text{Loc}_i)$
  \STATE $\Phi_i\leftarrow$SimulateCosts$\left(\{\textbf{r}^{m'}\}_{m'=1}^{M'}, \{\textbf{a}^{m'}\}_{m'=1}^{M'}\right)$
  \IF {$\Phi_i <$ SimulateCosts($\text{WT}_i$)}
  \STATE Replace$\left(\text{WT}_i, \{\textbf{r}^{m'}\}_{m'=1}^{M'}, \{\textbf{a}^{m'}\}_{m'=1}^{M'}\right)$
  \ENDIF
  \ENDFOR
 \end{algorithmic}
 \end{algorithm}
 \begin{figure}[!b]
    \centering
    \begin{subfigure}[b]{0.3\linewidth}
        \centering
        \includegraphics[width=\linewidth]{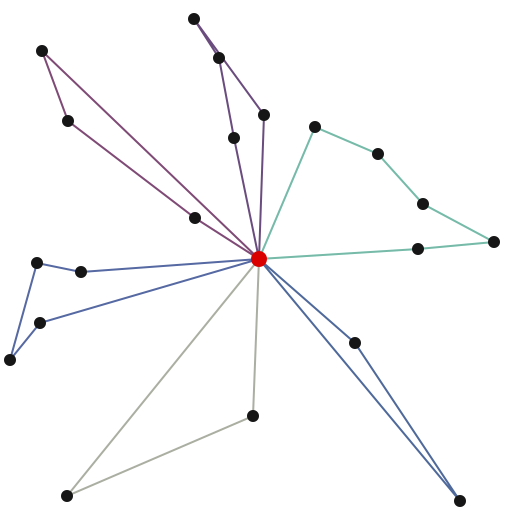}
        \caption[Routes]{{\small $p_C=0.01$}}    
        \label{fig:route_0.01_nometa}
    \end{subfigure}
    \begin{subfigure}[b]{0.3\linewidth}  
        \centering 
        \includegraphics[width=\linewidth]{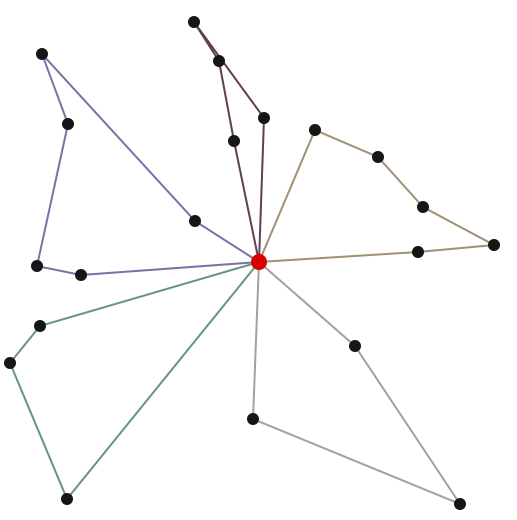}
        \caption[Routes]{{\small $p_C=0.1$}}    
        \label{fig:route_0.1_nometa}
    \end{subfigure}
    \begin{subfigure}[b]{0.3\linewidth}   
        \centering 
        \includegraphics[width=\linewidth]{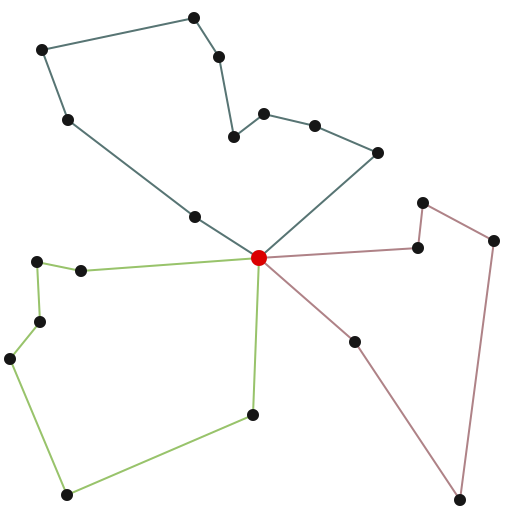}
        \caption[Routes]{{\small $p_C=0.5$}}    
        \label{fig:route_0.5_nometa}
    \end{subfigure}
    \caption{\small Routes for $N=20$ Customers without RF Metaheuristic} 
    \label{fig:routes_1}
    \hfill
    \hfill
    \hfill
    \hfill
    \begin{subfigure}[b]{0.3\linewidth}
        \centering
        \includegraphics[width=\linewidth]{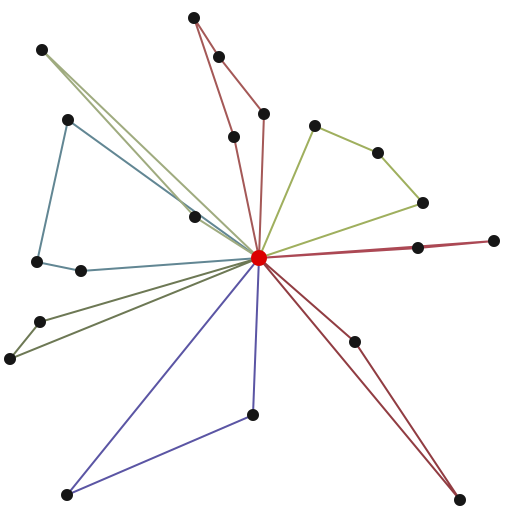}
        \caption[Routes]{{\small $p_C=0.01$}}    
        \label{fig:route_0.01_meta}
    \end{subfigure}
    \begin{subfigure}[b]{0.3\linewidth}  
        \centering 
        \includegraphics[width=\linewidth]{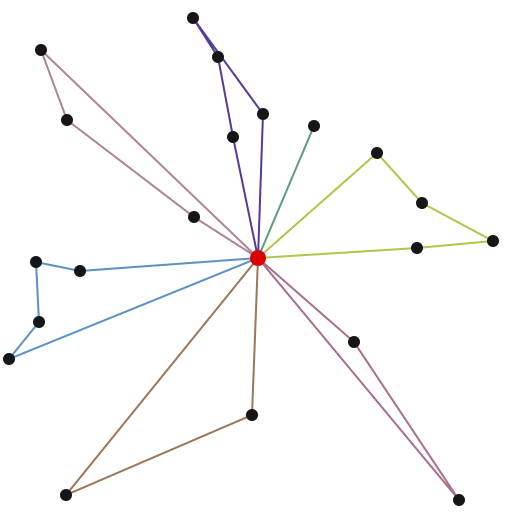}
        \caption[Routes]{{\small $p_C=0.1$}}    
        \label{fig:route_0.1_meta}
    \end{subfigure}
    \begin{subfigure}[b]{0.3\linewidth}   
        \centering 
        \includegraphics[width=\linewidth]{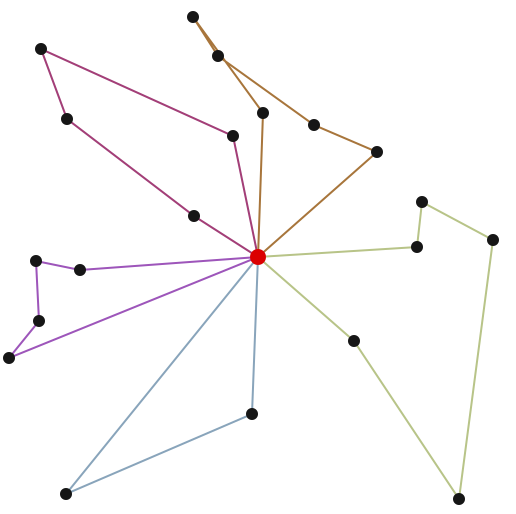}
        \caption[Routes]{{\small $p_C=0.5$}}    
        \label{fig:route_0.5_meta}
    \end{subfigure}
    \caption{\small Routes for $N=20$ Customers with RF Metaheuristic} 
    \label{fig:routes_2}
\end{figure}
The description of WorstSimulatedTeams and Replace is available in Appendix A. We begin by simulating the total cost $\Phi$ for given routes $\{\textbf{r}^m\}_{m=1}^M$ and appointment schedules $\{\textbf{a}^m\}_{m=1}^M$. Then for each $i=1,2,...,M$, we obtain the $i$ service teams with the lowest simulated total costs, and consider all customers that are a part of those service team's routes, giving us the subset $\text{WT}_i\subset [N]$. We then run the H-SARA Solver on $\text{WT}_i$ using the different routing and scheduling models, and choose the solution with the lowest simulated total cost. If this simulated total cost for $\text{WT}_i$ is lower than the original simulated cost for the $i$ worst service teams, we return a separate H-SARA Solution that copies the original solution but replaces the routes and schedules for the $i$ worst service teams with the new routes and schedules.
\begin{figure*}[!b]
\centering
    \begin{subfigure}{0.475\linewidth}
        \centering
        \includegraphics[width=\linewidth]{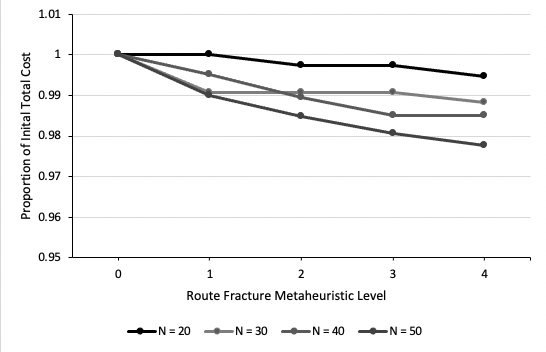}
        \caption[Routes]{{\small H-SARA-0 Improvement}}    
        \label{fig:H0_improve_meta}
    \end{subfigure}
    \begin{subfigure}{0.475\linewidth}  
        \centering 
        \includegraphics[width=\linewidth]{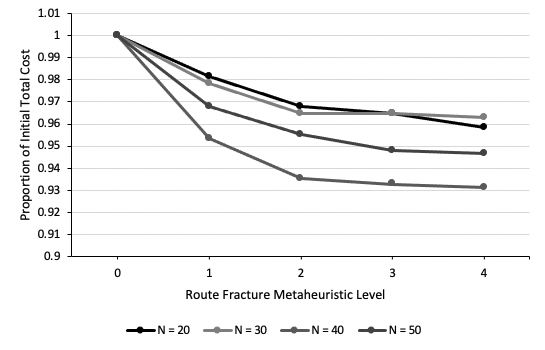}
        \caption[Routes]{{\small H-SARA-1 Improvement}}    
        \label{fig:H1_improve_meta}
    \end{subfigure}
    \caption{\small Proportion of Improvement for Metaheuristic Level for $N=20,30,40,50$ and $p_C=0.1$}
    \label{fig:improve_meta}
\end{figure*}
The Route Fracture (RF) Metaheuristic builds on the above algorithm by iteratively improving an H-SARA Solution. For a given H-SARA Solution with routes and schedules $\{\textbf{r}^m\}_{m=1}^M$, $\{\textbf{a}^m\}_{m=1}^M$, the metaheuristic uses the Route Fracture Algorithm to generate a solution with the $i$ service teams that have lowest simulated total costs replaced, for $i=1,\cdots, M$. This constitutes the first level of iteration. If a better solution was found, we run the metaheuristic again on the new solution. We repeat this till a specified level of iteration or until no better solutions have been found.

\section{Computational Results}
In this section, we present some examples of H-SARA Input instances and provide the results of Route Generation, Appointment Scheduling, Impact of Cancellation, and the Impact of the RF Metaheuristic. We split this section into parts discussing the following categories:
\begin{enumerate}
    \item The average number $M$ of service teams required for random instances with various customer counts.
    \item Presenting generated routes for different routing models, cancellation models, usage of the RF metatheuristic.
    \item Comparing the simulated Scheduling Cost $\Phi_S$ of the different appointment scheduling methods on routes of increasing length
    \item Quantifying the degree of improvement for the simulated Total Cost $\Phi$ based on number of levels of RF Metaheuristic application.
\end{enumerate}
We first establish the baseline H-SARA parameters we use to generate instances to evaluate our model on:
\noindent\makebox[\linewidth]{\rule{\linewidth}{1pt}}
\small $N\in\{20,30,40,50\}$ (Number of Customers)\\
$L = 480$ (Standard Service Hours)\\
$\mu_S =60, \sigma_S = 30$ (Mean and Variance of Service Time)\\
$t(i,j)=d(i,j)\xi/\exp(\sigma^2/2)$ (Distribution of Travel Times)\\
$f_m=250$ (Service Team Assignment Cost)\\
$\lambda_T=2$ (Unit Travel Cost)\\
$\lambda_W=10$ (Unit Customer Waiting Cost)\\
$\lambda_I = 5$ (Unit Service Team Idle Cost)\\
$\lambda_O = 15$ (Unit Service Team Overtime Cost)\\
$p_C\in\{0.01,0.1,0.5\}$ (Customer Service Cancellation Rate)\\
\noindent\makebox[\linewidth]{\rule{\linewidth}{1pt}}
\normalsize

These parameters are determined due to the following reasons. The end of service time $L$ is set to be $480$ minuets, i.e. $8$ hours, to reflect an $8$ hour work day. We fix the mean service team $\mu_S$, service team assignment cost, unit travel cost, unit customer waiting cost, unit service team idle cost, and unit service team overtime cost to the upper bounds of the Uniform distributions we set for the Case Study. Travel times between customers $i$ and $j$ takes on a scaled Log-Normal Distribution. In particular, the value of $\xi\sim \text{LnN}(\mu=0,\sigma = 0.5)$ to mimic the Bus Travel Time distributions for Sections as shown in Cats et al., (2014) \cite{travel_time_stops}

We examine the model's behavior for the number of customers $N=20,30,40,50$ since the value of $N$ directly affects the service team assignment and routing, and hence, the appointment scheduling. We further examine the behavior of the model under different customer cancellation rates $p_C=0.01,0.1,0.5$ representing ``almost no cancellation", ``medium cancellation", and ``high cancellation", respectively.
    
For each of these experiments we choose the value of the number of customers $N$ and the customer service cancellation rate $p_C$ as parameters. Using the specified value of $N$ and $p_C$, we uniformly generate $N$ customers in a $50\times 50$ km grid around the origin located at coordinate $(0,0)$. To simulate the routing and Scheduling Costs, we run the Monte Carlo Simulator for $500$ runs to obtain an empirical mean estimate of travel times, service times, waiting times, idle times, and overtime. This constitutes one trial. 
With the same H-SARA parameters, we then repeat the above trial process $50$ times for each experiment.
\begin{table}[!t]
  \centering
  \begin{tabular}{|z{0.08\linewidth}|z{0.13\linewidth}|z{0.13\linewidth}|z{0.13\linewidth}|z{0.13\linewidth}|z{0.13\linewidth}|}
    \hline
     \textbf{Rate} & \textbf{Model}  & \textbf{Waiting} & \textbf{Idling} & \textbf{Overtime} & \textbf{Cost} \\
    \hline
    \multirow{2}{*}{$0.01$}&Baseline & 11174.46 & \textbf{1718.63} & \textbf{885.54} & 13778.63\\
    &Simulated & \textbf{7468.25} & 2495.05 & 960.72 & \textbf{10924.02}\\
    \hline
    \multirow{2}{*}{$0.1$}&Baseline & 11117.44 & \textbf{1797.86} & \textbf{297.71} & 13213.01\\
    & Simulated & \textbf{7949.22} & 2562.99 & 462.79 & \textbf{10975.00}\\
    \hline
    \multirow{2}{*}{$0.5$}&Baseline & 9028.19 & \textbf{1427.72} & \textbf{95.89} & 10551.80 \\
    &Simulated & \textbf{6319.40} & 2377.33 & 106.70 & \textbf{8803.43}\\
    \hline
  \end{tabular}
  \caption{H-SARA-$0$ Scheduling Costs for $N=50$}
   \label{tab:hsara0-sched}
\hspace*{\fill}
  \begin{tabular}{|z{0.08\linewidth}|z{0.13\linewidth}|z{0.13\linewidth}|z{0.13\linewidth}|z{0.13\linewidth}|z{0.13\linewidth}|}
    \hline
     \textbf{Rate} & \textbf{Model}  & \textbf{Waiting} & \textbf{Idling} & \textbf{Overtime} & \textbf{Cost} \\
    \hline
    \multirow{2}{*}{$0.01$}&Baseline & 11523.78 & \textbf{1667.99} & \textbf{1052.04}& 14243.82\\
    &Simulated & \textbf{7914.37} & 2456.15 & 1387.68 & \textbf{11758.21} \\
    \hline
    \multirow{2}{*}{$0.1$}&Baseline & 11634.93 & \textbf{1711.03} & \textbf{646.52} & 13992.48\\
    & Simulated & \textbf{8156.42} & 2572.30 & 862.33 & \textbf{11591.05}\\
    \hline
    \multirow{2}{*}{$0.5$}&Baseline & 8927.38 & \textbf{1434.11} & 199.65 & 10561.14 \\
    &Simulated & \textbf{6817.22} & 2220.16 & \textbf{175.59} & \textbf{9212.97}\\
    \hline
  \end{tabular}
  \caption{H-SARA-$1$ Scheduling Costs for $N=50$}
\label{tab:hsara1-sched}
\end{table}
The techniques developed in Sections \hyperref[sec:routing]{\color{blue}VI}, \hyperref[sec:scheduling]{\color{blue}VII}, \hyperref[sec:cancellation]{\color{blue}VIII}, and \hyperref[sec:metaheuristic]{\color{blue}IX} were implemented in Python 3.7.9 using the following packages:
\begin{itemize}
    \item NumPy 1.20.1
    \item SciPy 1.6.1
    \item Google OR-Tools 8.2.8710.
\end{itemize} 
All experiments were performed on an Intel Core i5, with 2.133 GHz and 8 GB of RAM. 
\subsection{Service Team and Route Assignment}
Here we discuss the experimental results of the Service Team and Route Assignment portions of our model. Figure \ref{fig:M_nometa} shows the mean number of service teams $M$ to be assigned for $N=20,30,40,50$ customers, for different customer cancellation probabilities $p_c=0.01, 0.1, 0.5$ without the usage of the RF Metaheuristic, and Figure \ref{fig:M_meta} shows the mean number of service teams $M$ for the same instances using the RF Metaheuristic. These results are generated for H-SARA-$0$, since the route generation is the same for both H-SARA-$0$ and H-SARA-$1$. 

Following this, Figure \ref{fig:routes_1} contains examples of generated routes for $N=20$ customers in the H-SARA-$0$ cancellation model. We first fix certain locations for $N=20$ customers, and then generate routes for different customer cancellation probabilities $p_c=0.01, 0.1, 0.5$. Figure \ref{fig:routes_2} additionally shows the impact of the RF Metaheuristic on routes for these cancellation probabilities. Based on the above data, we note two major trends:
\begin{enumerate}
    \item Large cancellation probabilities results in a lower average number of service teams. This emerges due to one of our routing models considering expected customer service times as capacities. For higher cancellation probabilities, this expected service time decreases noticeably, and our CVRP routing model can produce teams that serve more customers on the route. This reduces the number of teams required to service all customers.
    \item The application of the RF Metaheuristic increases the number of teams assigned. This means that replacing the worst performing teams with additional teams to split the cost results in a better solution. We note that this increases the Routing Cost $\Phi_R$, but decreases the Scheduling Cost $\Phi_S$ enough to reduce the total cost $\Phi$. 
\end{enumerate}
\subsection{Customer Appointment Scheduling}
Here we compare the two different scheduling models of Baseline Appointment Times and Simulated Appointment Times and their impact on the Scheduling Cost $\Phi_S$, for $N=50$ and for customer cancellation probabilities $p_C=0.01,0.1,0.5$. Table \ref{tab:hsara0-sched} contains the results for H-SARA-$0$, and Table \ref{tab:hsara1-sched} contains the results similarly for H-SARA-$1$. We further represent the split of the Scheduling Cost into waiting, idle, and overtime cost. We do not use the RF Metaheuristic as we want to evaluate the scheduling model performance on initial solution states with longer routes. 

In both cases, we note that the Simulated Appointment Schedule result in a lower Scheduling Cost $\Phi_S$. However, this occurs primarily due to larger reductions in the Waiting Cost for the Simulated Appointment Times. The Baseline Appointment Schedule has lower Idle and Overtime Costs than the Simulated Schedules, but if the Unit Idle Cost $\lambda_I$ and the Unit Overtime Cost $\lambda_O$ vastly exceeds the Unit Wait Cost $\lambda_W$, then baseline schedules should result in lower simulated Scheduling Cost $\Phi_S$.

Finally, we note that higher cancellation rate results in lower Scheduling Cost. This follows since customer cancellation incurs no additional Wait Cost, Idle Cost, and Overtime Cost at a canceled customer in H-SARA-$0$, and even reduces Overtime for teams. In H-SARA-$1$, higher cancellation rate gives teams more opportunities to reroute, thereby further reducing the simulated Total Cost.
\subsection{Route Fracture Metaheuristic}
Finally in Figure \ref{fig:improve_meta} we evaluate the rate of improvement of the total cost $\Phi$ by the RF Metaheuristic for number of iterations. We present the proportional improvement of the total cost for the Customer Counts of $N=20,30,40,50$ with Cancellation Rate $p_C=0.1$ for $4$ iterations of the RF Metaheuristic. 

Figure \ref{fig:H0_improve_meta} presents this proportional improvement for H-SARA-$0$. We note that for lower Customer Counts such as $N=20,30$, the minimum Total Cost $\Phi$ either stagnates immediately or decreases slightly before stagnating again. This indicates that the RF Metaheuristic tends to stop after one iteration of improvement for smaller instances. For larger instances, the improvement is a lot more pronounced and tends to persist after numerous iterations. Finally, note that the proportion of improvement is more pronounced for earlier iterations indicating convergence to a local minima for Expected Total Cost $\Phi$.

Similarly, Figure \ref{fig:H1_improve_meta} presents this proportional improvement for H-SARA-$1$. We first note that the proportional improvement across all classes of instances is a lot more substantial than in H-SARA-$0$ (reaching an improvement of almost $7\%$ over the $2\%$ improvement in the H-SARA-$0$ case). Further, the improvement persists over multiple iterations of the RF Metaheuristic, but is still more pronounced for earlier iterations. Again, this seems to indicate convergence to a local minima for Expected Total Cost $\Phi$.

\section{Conclusions \& Future Work}

Based on the above results, we note that the Expected Total Costs $\Phi$ for instances depend more on Scheduling Costs $\Phi_S$ than on the Routing Costs $\Phi_R$. As such, the different scheduling models and the unit costs associated with scheduling have a large impact on the effectiveness of our solution. The Simulated Schedule Model functions better for instances where the Unit Wait Cost $\lambda_W$ is larger than the Unit Idle Cost $\lambda_I$, while the Baseline Schedule Model functions better for instances with much larger Unit Idle Costs. 

The impact of cancellation seems to directly affect the expected number of teams $M$ required. Higher rates of cancellation results in fewer teams required to service all customers. Further, higher cancellation rate results in lower Expected Scheduling Cost. Finally, the RF Metaheuristic increases the number of teams hired, and results in the most improvement in earlier iterations. 

A potential improvement to routing models would be a Stochastic VRP formulation with stochastic service times and travel times. Another potential improvement emphasizes the recursive form for Waiting Times, Idle Times, and Overtime in Eqn. \ref{eqn:sched}. Similar to how Gupta and Denton (2008) \cite{asp_gupta_denton} modelled the Appointment Scheduling Problem for a Single Stochastic Server as a $2$-Stage Stochastic Linear Program, Eqn. \ref{eqn:sched} could be used to construct a similar Stochastic Linear Program incorporating Travel Times, in addition to Service Times and Cancellation Rates. This would result in near optimal appointment scheduling for single teams in the H-SARA-$0$ Model of Cancellation. 

The H-SARA-$1$ Model of Cancellation requires a better formulation of the routes, waiting times, idle times, and overtime in order to express dynamic team rerouting better. However, we do believe that the additional flexibility offered by Customer Cancellation Time Distributions results in a more realistic Home Service model, with good scope for improvement in formulation and methods. Additionally, the generalization of the models of cancellation into H-SARA-$\lambda$ for $\lambda\in[0,1]$ gives more flexibility in the structure of cancellation and can incorporate a blend of ``forgetful" and ``responsible" customers.

\bibliographystyle{IEEEtran}
\bibliography{IEEEabrv, biblio}
\appendices
\section{Miscellaneous Algorithms}
Here we discuss the implementation of our Simulation Algorithm that is used in Algorithm 1 (Monte Carlo Scheduler) to simulate Arrival times, Algorithm 2 (H-SARA-1 Reroute Heuristic) to simulate Scheduling, Routing, and Total Costs, and in Algorithm 3 (Route Fracture Algorithm) to determine the worst performing teams.

Suppose we have service team routes $\{\textbf{r}^m\}_{m=1}^M$, customer schedules $\{\textbf{a}^m\}_{m=1}^M$, and standard service hours $[0,L]$. For a service team $m$, with route $\textbf{r}^m$ and customer schedules $\textbf{a}^m$, we begin at the office with index $0$ and set the current time $\mathcal{T}=0$. The team first visits $r^m_1$, and so we sample from $\mathcal{D}_{T(r^m_0, r^m_1)}$ to determine the travel time $t(r^m_0, r^m_1)$ to customer $r^m_1$. This incurs a travel cost with Unit Travel Cost $\lambda_T$. Upon reaching customer $r^m_1$ at time $\mathcal{T} = t(r^m_0, r^m_1)$, if $\mathcal{T} < a^m_1$m the team then shifts into an ``idle" state  until service starts (incurring idle cost with Unit Idle Cost $\lambda_I$). Otherwise the service team $m$ enters a ``service" state where the service time $\mathcal{Z}^m_i$ is sampled from $\mathcal{D}_S$. If it shifts into a service state and $\mathcal{T} > a^m_1$, we incur waiting cost with Unit Wait Cost $\lambda_I$. Upon completion of service, the team departs for customer $r^m_2$. We repeat the above process till the team reaches the office at time $\mathcal{T}'$. If $\mathcal{T}'>L$ it incurs an overtime cost with Unit Overtime Cost $\lambda_O$. This can be more formally described as follows:

 \begin{algorithm}[H]
 \caption{Simulation Algorithm}
 \begin{algorithmic}[1]
 \renewcommand{\algorithmicrequire}{\textbf{Input:}}
 \REQUIRE $\textbf{r}^m$, $\textbf{a}^m$, $L$, $\lambda_T$, $\lambda_W$, $\lambda_I$, $\lambda_O$
 \STATE $\Phi^m_T=0$, $\Phi^m_I=0$, $\Phi^m_W=0$
 \STATE $\mathcal{T} = 0$
  \FOR {$i = 0$ to $n_m$}
  \STATE Sample $t(r^m_i, r^m_{i+1})$ from $\mathcal{D}_{T(r^m_i, r^m_{i+1})}$
  \STATE $\Phi^m_T\leftarrow \Phi^m_T + \lambda_Tt(r^m_i, r^m_{i+1})$
  \STATE $\mathcal{T}\leftarrow \mathcal{T} + t(r^m_i, r^m_{i+1})$
  \IF {$\mathcal{T} < a^m_{i+1}$}
  \STATE $\Phi^m_I\leftarrow \Phi^m_I + \lambda_I(a^m_{i+1}-\mathcal{T})$
  \STATE $\mathcal{T} = a^m_{i+1}$
  \ENDIF
  \STATE $\Phi^m_W\leftarrow \Phi^m_W + \lambda_W(\mathcal{T} - a^m_{i+1})^+$
  \STATE Sample $\mathcal{Z}^m_i$ from $\mathcal{D}_S$
  \STATE $\mathcal{T}\leftarrow \mathcal{T} + \mathcal{Z}^m_i$
  \ENDFOR
  \STATE $\Phi^m_O = \lambda_O(\mathcal{T}- L)^+$
  \RETURN $\Phi^m_T$, $\Phi^m_I$, $\Phi^m_W$, $\Phi^m_O$
 \end{algorithmic}
 \end{algorithm}

In Algorithm 1, we simply store the values of $\mathcal{T}$ whenever it arrives at customer $r^m_i$ to simulate Arrival times. In Algorithm 2 to simulate the Scheduling Cost, Routing Cost, and Total Cost, we run the above on a subsets $\{r^m_i, r^m_{i+1}, r^m_{i+2}\}$, $\{a^m_i, a^m_{i+1}, a^m_{i+2}\}$, and $\{r^m_i, r^m_{i+2}\}$, $\{a^m_i, a^m_{i+2}\}$. In Algorithm 3, since the simulation provides the Travel Cost, Wait Cost, Idle Cost, and Overtime Cost for a service team $m$, we can obtain the worst performing service teams to reroute and reschedule.

Additionally, Algorithm 3 uses a Replace subroutine that takes the nodes of the $i$ worst performing service teams and removes those $i$ service teams and their routes. It then replaces them with $M'$ service teams that have routes $\{\textbf{r}^{m'}\}_{m'=1}^{M'}$, and customer schedules $\{\textbf{a}^{m'}\}_{m'=1}^{M'}$.

\section{Graphical User Interface Tutorial}
\begin{figure*}[!t]
    \centering
    \includegraphics[width=\linewidth]{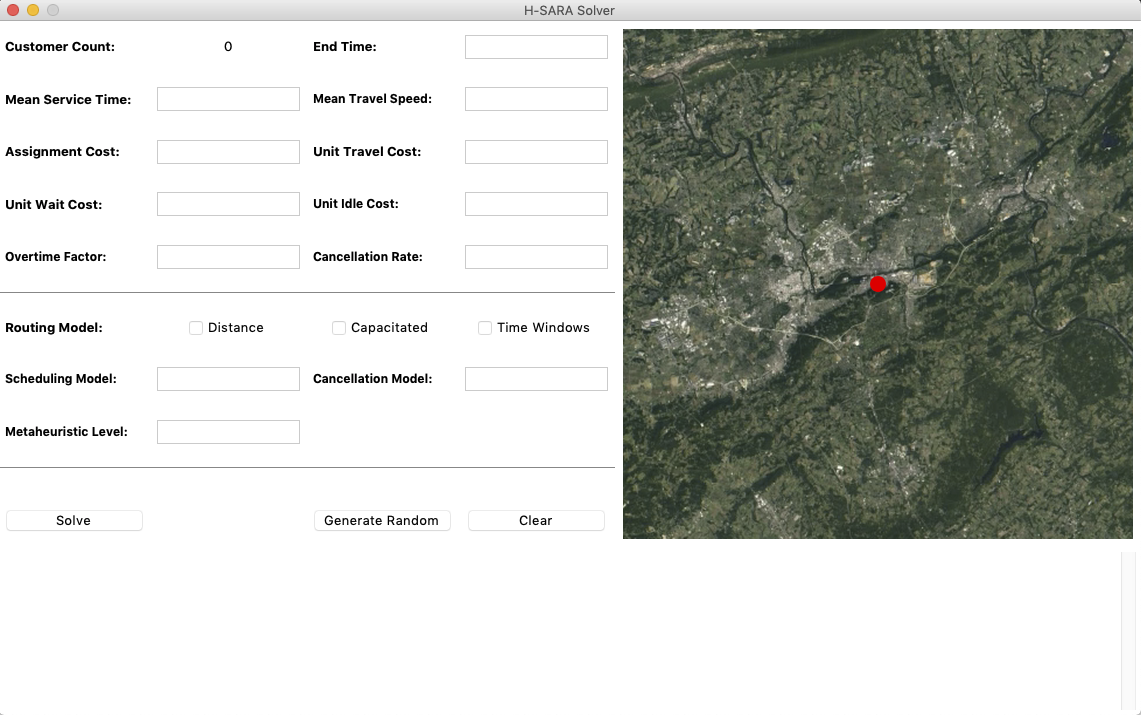}    
    \caption{\small H-SARA Solver UI} 
    \label{fig:hsara_ui}
\end{figure*}

The following is a brief overview on how to run the provided Graphical User Interface (GUI). As mentioned above, this model was implemented in Python 3.7.9 using the following packages:
\begin{itemize}
    \item NumPy 1.20.1
    \item SciPy 1.6.1
    \item Google OR-Tools 8.2.8710.
\end{itemize}
In order to run the GUI, please use the following command in terminal:
\[\texttt{python main.py --run\_style=ui}\]
This will bring up the screen shown in Figure \ref{fig:hsara_ui}.
The GUI has the following features:
\begin{enumerate}
    \item\textbf{Interactable Map:} The map available is a $50\times 50$km square centered around Lehigh University's coordinates $40.6049^{\circ} N, 75.3775^{\circ} W$. The red dot at the center represents the service office or the depot located at the origin. To add customers, click anywhere on the map! This will add a new customer to the list along with their location relative to the depot.
    \item\textbf{H-SARA Case Study Parameters: } The entry boxes in the top-left are used to input custom parameters for the H-SARA Problem. We have the following parameters:
    \begin{itemize}
        \item \textbf{Customer Count:} This corresponds to the number of customers and their locations to solve the H-SARA Problem. This can be modified by clicking on the interactable map.
        \item \textbf{End Time:} This represents the end of the standard service hours $[0, L]$ for all the service teams.
        \item \textbf{Mean Service Time:} This value specifies the mean $\mu_S$ of the service time distribution $\mathcal{D}_S$. We further set the standard deviation $\sigma_S$ of $\mathcal{D}_S$ to be $0.5\mu_S$. The service time distribution is the Uniform distrbution with mean $\mathcal{D}_S$ and variance $\sigma_S^2$
        \item \textbf{Mean Travel Speed:} This specifies the mean of the travel speed $\mu_V$, and directly affects the mean for all travel time distributions $\mathcal{D}_{T(i,j)}$, by setting the mean $\mu_{T(i,j)}=d(i,j)/\mu_V$. The travel time $t(i,j)=\mu_{T(i,j)}\xi/\exp(\sigma^2/2)$ where $\xi$ has the Log-Normal distribution $\text{LnN}(\mu=0,\sigma = 0.5)$
        \item \textbf{Assignment Cost:} This specifies the assignment cost $f_m$ for each of the $M$ homogenous service teams, as part of the Routing Cost $\Phi_R$
        \item \textbf{Unit Travel Cost:} This specifies the unit travel cost $\lambda_T$ that scales the service team travel time as part of the Routing Cost $\Phi_R$.
        \item \textbf{Unit Wait Cost:} This specifies the unit travel cost $\lambda_W$ that scales the customer waiting time when a service team arrives after the customer's scheduled time as part of the Scheduling Cost $\Phi_S$.
        \item \textbf{Unit Idle Cost:} This specifies the unit travel cost $\lambda_I$ that scales the service team idle time when a service team arrives before the customer's scheduled time as part of the Scheduling Cost $\Phi_S$.
        \item \textbf{Overtime Factor:} This specifies the unit overtime cost $\lambda_O$ that scales all time spent by the service team after their standard service hours $[0,L]$, as part of the Scheduling Cost $\Phi_S$.
        \item \textbf{Cancellation Rate:} This parameter specifies the probability $p_C$ a customer will cancel their requested service during the day. For H-SARA-$1$, the cancellation time distribution for a customer $r^m_i$ is $U[0, a^m_i]$.
    \end{itemize}
    \item\textbf{Model Hyperparameters:} The entry boxes in the center-left are used to choose various model changing parameters that can be modified to see the changes made to the solution. We have the following hyperparameters: \balance
    \begin{itemize}
        \item \textbf{Routing Model:} This changes the underlying VRP Solver constraints when generating routes. The options available the following:
        \begin{enumerate}
            \item \textbf{Distance:} This corresponds to a VRP Solver that tries to minimize the overall distance using $M$ teams, for $M=1,2\cdots, \lfloor N/3 \rfloor$
            \item \textbf{Capacity:} This corresponds to a CVRP Solver that tries to minimize the overall distance using $M$ teams, for $M=1,2\cdots, \lfloor N/3 \rfloor$
            \item \textbf{Time Windows:} This corresponds to a VRPTW Solver that tries to minimize the overall distance using $M$ teams, for $M=1,2\cdots, \lfloor N/3 \rfloor$
        \end{enumerate}
        \item \textbf{Scheduling Model:} This specifies the Scheduling Model to be used to generated schedules for all routes. Setting this hyperparameter to $1$ results in the Baseline Scheduling Model, setting it to $2$ results in the Simulated Scheduling Model, and setting it to $3$ results in producing solutions from each model.
        \item \textbf{Cancellation Model:} This specifies the values of $\lambda$ for H-SARA-$\lambda$. Setting this hyperparameter to $0$ gives us the H-SARA-$0$ problem variant where customer cancellation follows a ``last-minute" model, and setting this to $1$ gives us the H-SARA-$1$ problem variant where customer cancellation follows a model where customers notify of their cancellation at some random time between the start of the day and their appointment time (follows a uniform distribution).
        \item \textbf{Metaheuristic Level:} This specifies the maximum number of levels to run the RF Metaheuristic for. Setting it to $0$ results in no usage of the RF Metaheuristic. Setting it an integer greater than $0$ lets the RF Metaheuristic run till that level of iteration, or until it cannot find a better solution.
    \end{itemize}
\end{enumerate}

We further have three important buttons on the bottom:
\begin{enumerate}
    \item \textbf{Solve:} This generates the solution to the H-SARA Problem generated from the specified parameters, using the model specified by the hyperparameters. To get the best solution to H-SARA, we recommend selecting all options on the \textbf{Routing Model} section, setting \textbf{Scheduling Model} to $3$, and setting the \textbf{Metaheuristic Level} to $3$.
    \item \textbf{Generate Random: } This generates a random H-SARA Case Study instance from the following parameters:
    \begin{itemize}
        \item $N\in\{20,30,40, 50\}$
        \item $L\in \{240, 480, 720, 1200\}$
        \item $\mu_S\sim U[30,60]$
        \item $\mu_V = 1 \implies \mu_{T(i,j)}=d(i,j)$
        \item $f_m\in U[100,250]$
        \item $\lambda_T\in \{0.5,1,2\}$
        \item $\lambda_W\sim U[0,10]$
        \item $\lambda_I\sim U[0,5]$
        \item $\lambda_O\sim U[10, 15]$
        \item $p_C\in\{0.01, 0.05, 0.1\}$
        \item A non-empty subset of $\{\text{Distance}$, $\text{Capacitated}$, $\text{Time Windows}\}$ for the Routing Model
        \item Random Scheduling Model from Baseline (1), Scheduling (2), or Both (3)
        \item Random $\lambda=0.0$ or $\lambda=1.0$ for the Cancellation Model H-SARA-$\lambda$
        \item Random maximum Metaheuristic Level of $3$
    \end{itemize}
    \item \textbf{Clear: } This clears the map and all parameter and hyperparameter settings, allowing the testing of a different instance.
\end{enumerate}
On the bottom of the screen, below the buttons and the map, we have a text box that prints the Routes, Schedules, and simulates all of the costs.
\end{document}